\documentclass[10pt,twocolumn,letterpaper]{article}

\usepackage{iccv}              
\usepackage{multirow}
\usepackage[pagebackref,breaklinks,colorlinks,allcolors=iccvblue]{hyperref}

\usepackage{booktabs}
\definecolor{iccvblue}{rgb}{0.21,0.49,0.74}
\usepackage[pagebackref,breaklinks,colorlinks,allcolors=iccvblue]{hyperref}

\def\paperID{14547} 
\def\confName{ICCV}
\def\confYear{2025}

\title{RoCo-Sim: Enhancing Roadside Collaborative Perception through \\ Foreground Simulation}

\author{Yuwen Du$^{1,2}$\thanks{Equal contribution.} \quad Anning Hu$^{1}\footnotemark[1]$  \quad Zichen Chao$^{3}$ \quad Yifan Lu$^{1}$ \quad Junhao Ge$^{1}$ \quad Genjia Liu$^{1}$ \\
\quad Weitao Wu$^{3}$ \quad Lanjun Wang$^{2}$ \quad Siheng Chen$^{1\dagger}$ \\
\small$^{1}$ Shanghai Jiao Tong University \quad $^{2}$ Tianjin University \quad $^{3}$ Nanjing University of Science and Technology \\
\small \texttt{\{huanning, yifan\_lu, cancaries, LGJ1zed, sihengc\}@sjtu.edu.cn} \\
\small \texttt{\{duyuwen,wanglanjun\}@tju.edu.cn}, \texttt{\{zichen.chao, weitaowwtw\}@njust.edu.cn}
}

\date{} 

\begin{document}
\maketitle
\begin{abstract}
Roadside Collaborative Perception refers to a system where multiple roadside units collaborate to pool their perceptual data, assisting vehicles in enhancing their environmental awareness. Existing roadside perception methods concentrate on model design but overlook data issues like calibration errors, sparse information, and multi-view consistency, leading to poor performance on recent published datasets. 
To significantly enhance roadside collaborative perception and address critical data issues, 
we present the first simulation framework RoCo-Sim for road-side collaborative perception. RoCo-Sim is capable of generating diverse, multi-view consistent simulated roadside data through dynamic foreground editing and full-scene style transfer of a single image. RoCo-Sim consists of four components: (1) Camera Extrinsic Optimizer ensures accurate 3D to 2D projection for roadside cameras; (2) A novel Multi-View Occlusion-Aware Sampler (MOAS) determines the placement of diverse digital assets within 3D space; (3) DepthSAM innovatively models foreground-background relationships from single-frame fixed-view images, ensuring multi-view consistency of foreground; and (4) Scalable Post-Processing Toolkit generates more realistic and enriched scenes through style transfer and other enhancements. RoCo-Sim significantly improves roadside 3D object detection, outperforming SOTA methods by \textbf{83.74\%} on Rcooper-Intersection and \textbf{83.12\%} on TUMTraf-V2X for AP70. RoCo-Sim fills a critical gap in roadside perception simulation. Code will be released soon: \href{https://github.com/duyuwen-duen/RoCo-Sim}{https://github.com/duyuwen-duen/RoCo-Sim}

\end{abstract}
\begin{figure}[ht]
    \centering
    \includegraphics[width=0.8\linewidth]{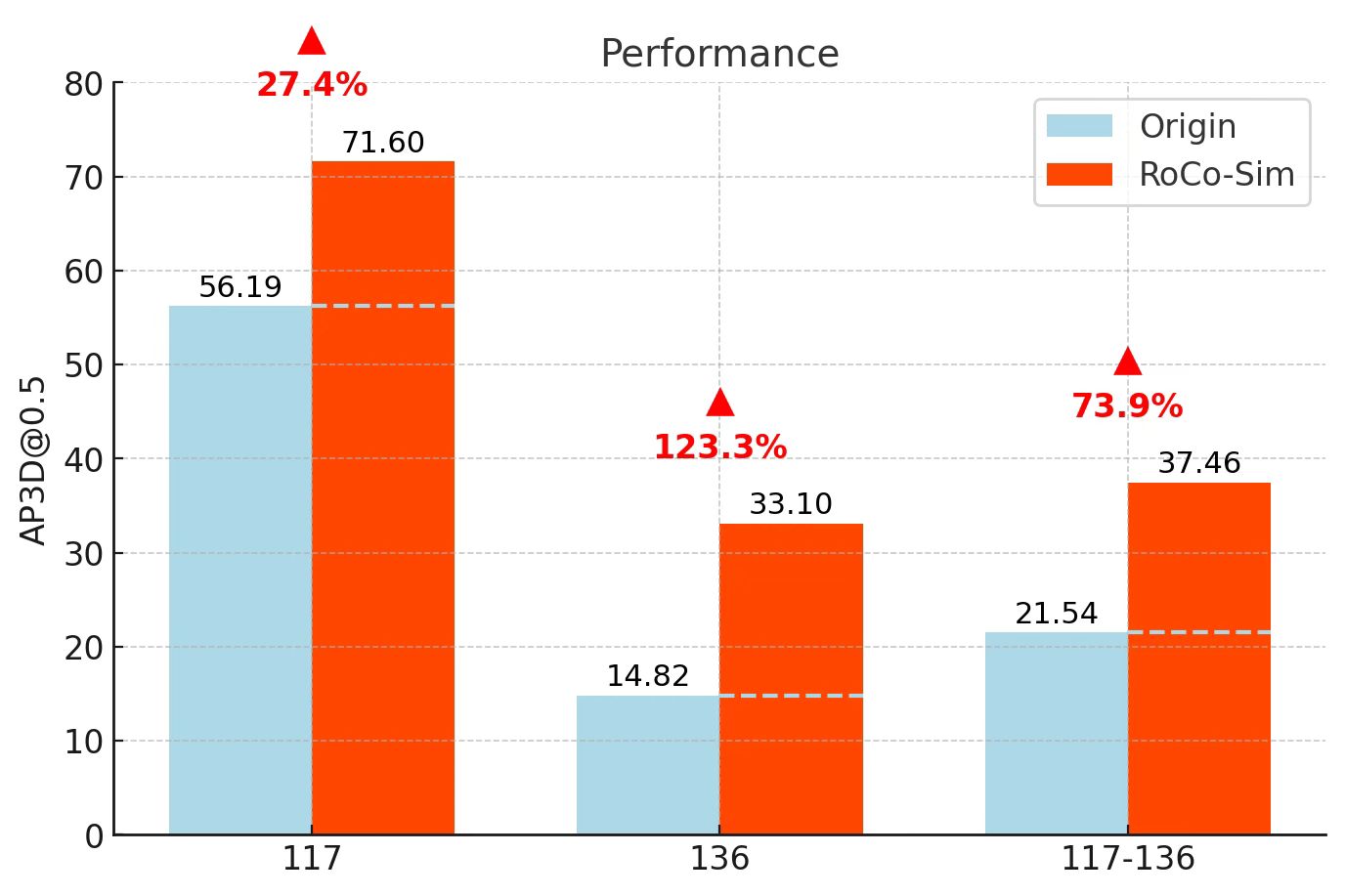}
    \caption{\textbf{Performance on RCooper-Intersection\cite{rcooper}, 117 and 136.} With RoCo-Sim, camera-only 3D detection model BEVHeight~\cite{yang2023bevheight}  achieves significantly better performance.}
    \label{fig:bar_plot}
    \vspace{-0.7cm}
\end{figure}

\begin{figure*}[!h]
\centering
\includegraphics[width=0.95\linewidth]{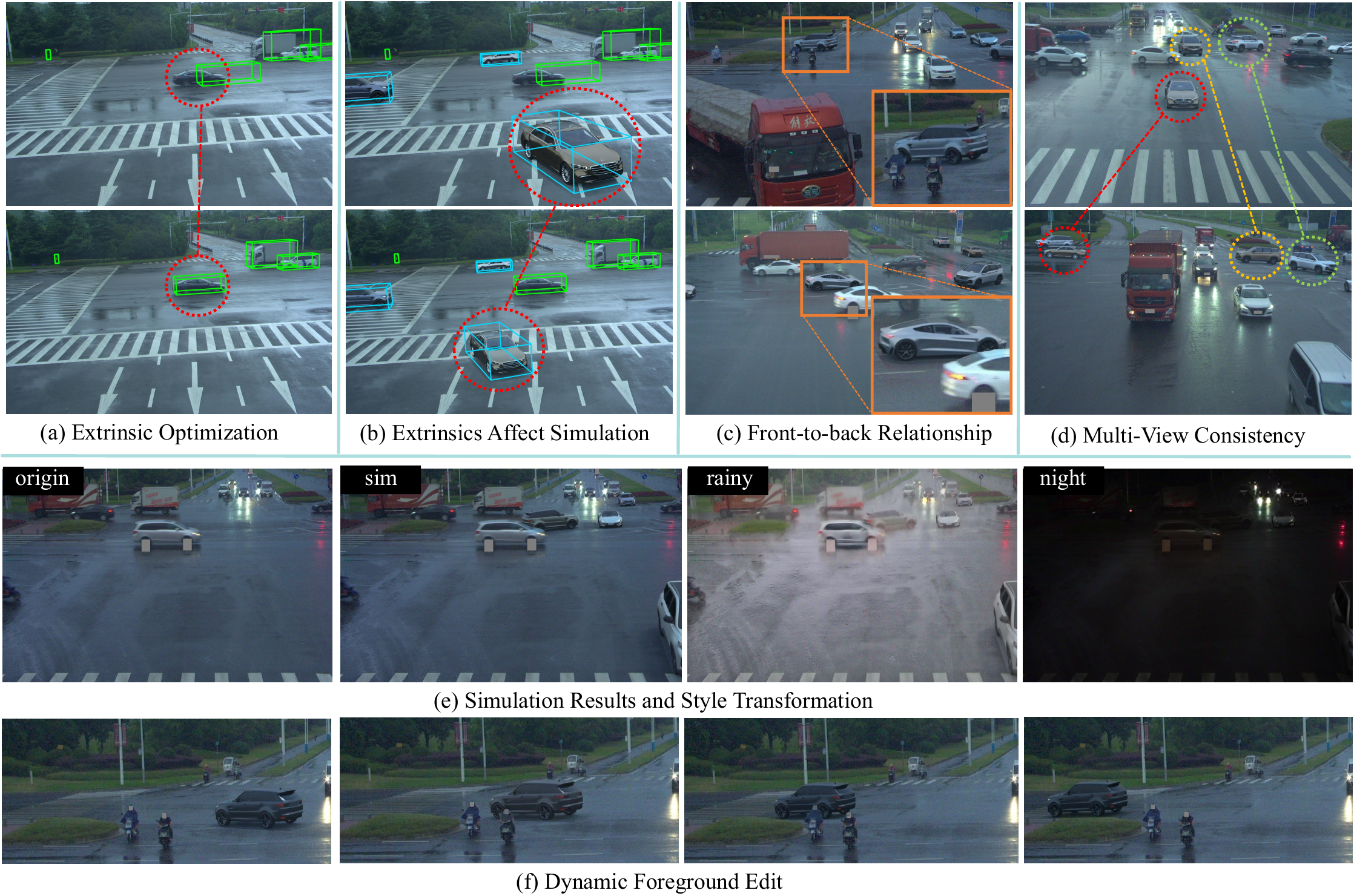}
   \caption{(a)  After extrinsic optimization, the projection of 3D bounding boxes is more accurate. (b) Camera extrinsic calibration can ensure the correct imaging of simulated objects, reducing the gap between simulated and real objects. (c) DepthSAM allows 3D simulated objects to be rendered onto 2D images while preserving the front-to-back relationships between objects. (d) Roco-Sim simulation can ensure multi-view consistency. (e) Roco-Sim can generate realistic and diverse simulated scenes. ``origin" refers to the original image, and ``sim" depicts the simulation of 4 cars. Furthermore, with the tools from the Post-Processing Toolkit, a variety of rich simulated scenarios can be created, such as rainy and night scenes. (f) RoCo-Sim can dynamically edit the foreground to generate simulated scenes with trajectory-based foreground objects, accurately maintaining front-to-back spatial relationships. As shown in the image, the simulated vehicle is occluded by the motorcycle when passing it, and it correctly generates vehicle shadows during movement.}
   \label{fig:Visual Results of the Simulation}
   \vspace{-0.3cm}
\end{figure*}   
\vspace{-0.3cm}
\section{Introduction}
\label{sec:intro}
\hspace{1em}Roadside Collaborative Perception ~\cite{yang2023bevheight,wang2024bevspread,monoGAE,monoground}  is a cooperative system that integrates perceptual data from multiple roadside units to help vehicles improve their understanding of the surrounding environment. This system serves as an effective supplement to autonomous driving, enhancing vehicle safety by addressing blind spots and providing a more comprehensive view of the road conditions. In Roadside Collaborative Perception, camera-based systems play a pivotal role by capturing rich visual information that can be leveraged to infer detailed 3D environmental characteristics. However, current methods utilizing camera-only systems for 3D detection in roadside collaborative perception have not yet achieved optimal results. \par


Existing roadside perception methods~\cite{li2023bevdepth,wang2024bevspread,monoGAE,monoground}, primarily focus on designing model architectures that aim to capture more information from the available data. 
However, these methods often overlook data issues. Firstly, the extrinsics of fixed-view cameras are difficult to calibrate initially and tend to drift over time due to environmental factors like mounting shifts or external interference, requiring frequent recalibration. Secondly, during data collection, long periods with few or no vehicles in the field of view result in sparse information density. Thirdly, ensuring multi-view consistency in annotations is highly challenging, leading to low-quality labeled data and high data collection costs. These data issues limit the performance of existing methods, causing poor performance on recently published datasets\cite{rcooper,tum}.

Here, we highlight the key novelty of our approach. Unlike previous methods that focus on tweaking model architectures or laboriously annotating large datasets, we unleash the power of large-scale information-dense simulated data to dramatically enhance the performance of roadside perception models. Achieving this goal imposes several requirements on roadside simulation: simulations must be conducted using fixed-perspective inputs. The capability to edit scenes by inserting 3D foreground objects is essential for generating extensive simulated data. Then, the simulation must exhibit strong generalization capabilities, enabling deployment to new scenarios without extensive retraining, and produce photorealistic images.

However, many existing simulation methods are unsuitable for roadside perception tasks due to two key challenges: simulation with fixed-viewpoints and lack of 3D layouts for editing. As shown in Tab.\ref{tab:method campare}, methods based on NeRF~\cite{neurad,Chatsim,s-nerf,s-nerf++,unisim,mars} and 3DGS~\cite{Chatsim,omnire}  cannot reconstruct under the conditions of roadside fixed sparse views, making them ineffective. Moreover, most methods\cite{bevcontrol,bevgen,DrivingDiffusion,drivedreamer-2,DriveDreamer,Drive-WM},   that aim to generate simulated objects with 3D information rely on 3D layout inputs, which are lacking in roadside datasets. Traditional graphics engines like CARLA~\cite{carla} manually simulate scenes without multi-view images but are time-consuming, labor-intensive, and lack realism, making them unsuitable for roadside scenarios. Furthermore, deploying roadside collaborative perception in real-world applications requires simulation methods that can generate large-scale data across diverse road scenarios without additional scene-specific training.

To address these challenges, we introduce RoCo-Sim, the first simulation framework designed for roadside collaborative perception. RoCo-Sim is capable of generating diverse, multi-view consistent simulated roadside data through dynamic foreground editing and full-scene style transfer of a single image, as shown in Fig. \ref{fig:Visual Results of the Simulation}. Since the editing is performed in 3D space, our modifications made from one viewpoint can automatically propagate to other viewpoints, ensuring spatial consistency across multiple perspectives. The key design rationale behind RoCo-Sim is to leverage a rich 3D asset library to establish a 3D-to-2D mapping and, with the aid of graphical tools\cite{blender}, seamlessly render foreground objects onto real 2D backgrounds. RoCo-Sim encompasses 4 key components: 
(i) Camera Extrinsics Optimizer for accurate camera pose modeling; 
(ii) A novel Multi-View Occlusion-Aware Sampler (MOAS), which determines the occlusion-aware placement of digital assets within the 3D space, ensuring that assets are dynamically positioned in a physically plausible manner;
(iii) DepthSAM, which innovatively models the foreground-background relationships in each viewpoint, enabling occlusion-aware insertion of 3D assets while maintaining multi-view consistency; and (iv) Scalable Post-Processing Toolkit for seamless integration through style transfer and other enhancements. 


\vspace{0pt}
Extensive experiments are conducted on two real-world roadside datasets, RCooper~\cite{rcooper} and TUMTraf-V2X~\cite{tum}, to evaluate RoCo-Sim. Our findings indicate that: i) Camera Extrinsics Optimizer greatly improves model performance. Performance on RCooper improves by \textbf{62.55\%} for AP70. ii) The performance improvement for perception becomes more significant as the amount of simulation data and the number of simulated vehicles per image increase. iii) RoCo-Sim significantly enhances perception performance. It enhances previous SOTA performance by \textbf{83.74\%} on Rcooper-Intersection and \textbf{83.12\%} on TUMTraf-V2X for AP70. To sum up, our contributions are:
\begin{itemize}
     \item We propose the first simulation framework for roadside collaborative perception, which simulates 3D foreground objects and renders them onto real 2D backgrounds.
    
     \item RoCo-Sim is a structured modular framework designed to generate diverse, multi-view consistent roadside simulation data through dynamic foreground editing and full-scene style transfer, while ensuring occlusion-aware asset placement and seamless integration for realistic and scalable simulation.
    
    \item We conduct extensive experiments on two real-world roadside datasets, validating that RoCo-Sim greatly enhances model performance, surpassing the impact of algorithmic improvements.
\end{itemize}

\section{Related Work}
\textbf{Camera-only 3D object detection. } 
Camera-based methods recover 3D information from 2D images and detect objects in a shared BEV feature space, either implicitly or explicitly. Frustum-based methods~\cite{lss,li2023bevdepth} predict depth distributions to form 3D frustums, then apply voxel pooling~\cite{li2023bevstereo,reading2021categorical} to extract BEV features. Query-based methods~\cite{bevformer,wang2022detr3d,liu2023petrv2} use transformers to directly index 2D image features with 3D object queries. Recent approaches~\cite{streampetr} incorporate streaming images to better utilize spatiotemporal information. While most methods focus on vision-centric BEV detection, roadside 3D object detection adapts frustum-based methods by estimating height~\cite{yang2023bevheight} instead of depth. These methods focus on how to better extract spatiotemporal information from existing datasets.



\noindent\textbf{Collaborative perception. } Collaborative perception leverages multiple agents' sensors to provide broader coverage with fewer blind spots~\cite{hu2022where2comm}. Based on the stage of information transmission~\cite{heal}, it can be categorized into i) early fusion, which shares raw sensor data for complete perception but requires high bandwidth, ii) late fusion, which combines detected 3D boxes to save bandwidth but loses environmental details, and iii) intermediate fusion, which transmits low-resolution feature maps to balance performance and bandwidth~\cite{wei2024asynchrony,hu2023collaboration}. Existing systems mainly operate between vehicles or with a single roadside infrastructure~\cite{yang2024monogae,liu2024towards}, while camera-only roadside collaborative perception remains unexplored due to data collection and calibration challenges. Our simulator addresses these challenges to advance this field.

\begin{table}[!t]
    \caption{Comparison of existing methods for autonomous driving simulation. ``Fixed sparse views" refers to a setup where camera angles are fixed and limited in number. ``Zero-shot" means the ability to perform without training on specific scenes.}
    \centering
\resizebox{\columnwidth}{!}{

    \begin{tabular}{cccccc}
    \toprule
        Method & Photo-realistic & Dim. & Editable & Fixed sparse views & Zero-shot\\
        \midrule
       \midrule
        CARLA \cite{carla} & × & 3D & \checkmark & \checkmark & \checkmark\\
        BEVGen \cite{bevgen} & \checkmark & 2D & \checkmark & × & × \\
        BEVControl \cite{bevcontrol} & \checkmark & 2D  & \checkmark & × & × \\
        DriveDreamer \cite{DriveDreamer} & \checkmark & 2D &  \checkmark & × & × \\
        DriveDreamer-2 \cite{drivedreamer-2} & \checkmark & 2D &  \checkmark & × & × \\
        DrivingDiffusion \cite{DrivingDiffusion} & \checkmark & 2D & \checkmark & × & × \\
        MagicDrive \cite{MagicDrive} & \checkmark & 2D & \checkmark & × & × \\
         MagicDrive3D \cite{magicdrive3d} & \checkmark & 2D & \checkmark & × & × \\
        S-NeRF \cite{s-nerf} & \checkmark & 3D& × & × & × \\
        S-NeRF++ \cite{s-nerf++} & \checkmark & 3D& \checkmark & × & × \\
        UniSim \cite{unisim} & \checkmark & 3D & \checkmark & × & × \\
        MARS \cite{mars} & \checkmark & 3D & \checkmark & × & ×\\
        ChatSim \cite{Chatsim} & \checkmark & 3D & \checkmark & × & × \\
        OmniRe \cite{omnire} & \checkmark & 3D & \checkmark & × & × \\
        \textbf{Roco-Sim(Ours)} & \checkmark & 3D-2D & \checkmark & \checkmark & \checkmark \\
    \bottomrule
    \end{tabular}}

    \label{tab:method campare}
\vspace{-0.6cm}
\end{table}

\noindent\textbf{Realistic driving scenario simulation. }
Current realistic scene simulation methods can be generally divided into two categories: image generation, and scene rendering. Some  image generation works create realistic scene images that mimic data from real vehicle cameras. By employing advanced techniques like diffusion models~\cite{DriveDreamer,DrivingDiffusion,Drive-WM,Panacea,subjectdrive}, variational autoencoders~\cite{bevgen}, they can produce synthetic images tailored to various conditional requirements. However, most of them cannot maintain view consistency ~\cite{bevcontrol,bevgen,DrivingDiffusion,MagicDrive,Panacea,DriveDreamer} or can only generate data without producing corresponding bbox annotations~\cite{ADriver-I,GAIA}.
Another line of work based on NeRF~\cite{Chatsim,magicdrive3d,mapnerf,neuroncap,neurad,s-nerf,unisim} or 3DGS~\cite{magicdrive3d,drivinggaussian} targets to reconstruct the 3D scene and thus synthesize the camera visual image from novel views. However, this is not applicable in roadside scenarios, as the roadside cameras are fixed, making it difficult to achieve 3D modeling with multi-view cameras.Instead we focus on using 3D assets, and conduct multi-view camera and vehicle modeling, rendering the vehicle onto the background image to achieve foreground simulation, which ensures multi-view consistency while generating annotated simulation data.

\begin{figure*}[!h]
\centering
\includegraphics[width=1\linewidth]{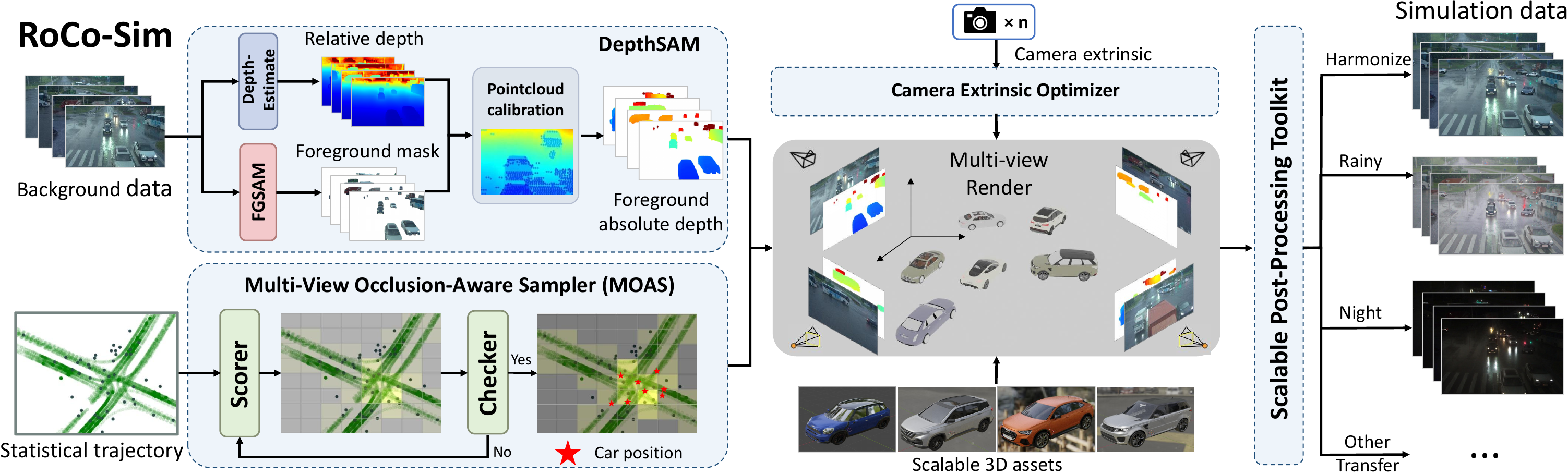}
   \caption{\textbf{Overview of RoCo-Sim.} RoCo-Sim consists of four components: (1) Camera Extrinsic Optimization ensures accurate roadside cameras modeling; (2) Multi-View Occlusion-Aware Sampler (MOAS) determines the placement of diverse digital assets within 3D space; (3) DepthSAM ensures that rendering adheres to front-to-back relationships and correct occlusion between objects; and (4) Scalable Post-Processing Toolkit is an expandable toolkit capable of generating more diverse and enriched scenes. }
   \label{fig:Framework}
   \vspace{-0.4cm}
\end{figure*}

\section{Roadside Collaborative Simulation}
\hspace{1em}As depicted in Fig.\ref{fig:Framework}, we introduce RoCo-Sim, the first simulation framework for roadside collaborative perception. It can generate diverse, multi-view consistent simulated roadside data from a single image and supports dynamic foreground editing and full-scene style transfer. Our key idea is to leverage a rich library of 3D assets to construct a 3D-to-2D mapping and render foreground objects onto real 2D backgrounds using graphical tools.\par

\subsection{Camera Extrinsic Optimizer}
\label{sec: Camera extrinsic optimizer}
\hspace{1em}Optimization of camera extrinsics can effectively reduce the calibration error of fixed roadside cameras, thereby facilitating the accurate operation of the simulation. By mathematically modeling the 3D to 2D projection process and using optimization algorithms to reduce the errors in camera extrinsics, we develop a convenient calibration tool that is applicable to various roadside cameras.

To refine the extrinsic parameters of roadside cameras, we optimize the transformation matrix \( \mathcal{M}_{\text{l2c}} = [\mathcal{R} | \mathcal{T}] \), where \( \mathcal{R} \in SO(3) \) is the rotation matrix and \( \mathcal{T} \in \mathbb{R}^{3 \times 1} \) is the translation vector. Due to calibration errors, the actual transformation deviates from an ideal estimate \( \mathcal{M}_{\text{l2c}}^{*} \), leading to an extrinsic error \( \Delta \mathcal{M} = [\Delta \mathcal{R} | \Delta \mathcal{T}] \), such that:
\begin{equation}
\mathcal{M}_{\text{l2c}} = \mathcal{M}_{\text{l2c}}^{*} + \Delta \mathcal{M}
\end{equation}

This error affects the 2D projection, introducing a discrepancy between the projected bounding box points \( B_{2d} = \mathcal{K} \cdot [\mathcal{R} | \mathcal{T}] \cdot B_{3d} \) and the manually adjusted keypoints \( B_{2d}^{\prime} \), where \( B_{3d} \) represents the 3D coordinates of the bounding box corner points. We directly solving for \( \Delta \mathcal{M} \) by minimizing the projection error:
\begin{equation}
\min_{\Delta \mathcal{R}, \Delta \mathcal{T}} ||\mathcal{K} \cdot ([\mathcal{R}^{*} + \Delta \mathcal{R} | \mathcal{T}^{*} + \Delta \mathcal{T}]) \cdot B_{3d} - B_{2d}^{\prime}||^{2}
\end{equation}subject to \( (\mathcal{R}^{*} + \Delta \mathcal{R}) (\mathcal{R}^{*} + \Delta \mathcal{R})^{T} = \mathcal{I} \) to ensure \( \mathcal{R} \) remains a valid rotation, where \(\mathcal{K}\in\mathbb{R}^{3 \times 3}\) is the camera intrinsic matrix. \(\mathcal{I}\in\mathbb{R}^{3 \times 3}\) represents the identity matrix. $\|\cdot\|^2$ represents L2 norm.  

We employ the Broyden–Fletcher–Goldfarb–Shanno (BFGS) optimization algorithm, a quasi-Newton method that iteratively approximates the inverse Hessian of the objective function to guide updates in \( \Delta \mathcal{R} \) and \( \Delta \mathcal{T} \). Unlike first-order gradient descent methods, BFGS accelerates convergence by incorporating second-order curvature information, making it well-suited for non-linear optimization problems. We leverage solvers such as SciPy for efficient computation.  

To further improve usability, we developed an interactive UI-based tool that allows users to manually adjust the projected bounding box, providing calibrated keypoints \( B_{2d}^{\prime} \), significantly reducing optimization complexity and improving efficiency. This tool will be released.

The optimized extrinsic matrix is then obtained as:

\begin{equation}
\label{equation3}
\mathcal{M}_{\text{l2c}}^{opt} = [\mathcal{R}^{opt} | \mathcal{T}^{opt}]
\end{equation}ensuring precise projection alignment and correctly imaging 3D assets to reduce the distribution gap between simulated foregrounds and real foregrounds.

\subsection{ Multi-View Occlusion-Aware Sampler}
\label{sec: Multi-View Occlusion-Aware Sampler}

\hspace{1em}To address the sparsity of existing data and the limitation that current methods require a 3D layout for editing, we introduce the Multi-View Occlusion-Aware Sampler (MOAS). MOAS automatically determines the placement of digital assets within a 3D space based on the distribution of objects in the scene. The placements must meet two criteria: they should be widely dispersed, and visible from multiple viewpoints while ensuring physical plausibility (e.g., no collisions). To achieve this, we sample potential placement points from locations where real objects have appeared and use a scorer to select sampling regions and a checker to detect insertability.


\textbf{Optimization-based scorer.}
 We divide the entire 3D space into multiple grids, score these grids to determine the selected placement areas, and randomly sample within them. The score is determined by the sum of two optimization objectives, where a higher score indicates better placement strategy.

One optimization objective is to maximize the visibility of placement positions from multiple perspectives while ensuring that these perspectives are as evenly distributed as possible; that is:
\begin{equation}
\operatorname{avg}=\frac{\sum_{m \in M}\left(V_{m}\left(p_{i}\right)\right)}{|M|}
\end{equation}
\begin{equation}
V_{i}=\sum_{m \in M} V_{m}\left(p_{i}\right)-\sum_{m \in M}\left(V_{m}\left(p_{i}\right)-\operatorname{avg}\right)^{2}
\end{equation}
where $M$ is the camera lists, ${|\cdot|}$ is the is the cardinal of set. $p_i$ represents the chosen insert points. $V_m(p_i)$ indicates whether $p_i$ is visible in camera $m$; if it is visible, the value is 1; otherwise, it is 0. $V_i$ measures the visibility and coverage of the field of view for the $i$-th placed point.\par
Another optimization objective is to maximize the distance between points, ensuring that the distribution of placed vehicles is as dispersed as possible, which creates a richer set of placement points. That is:
\begin{equation}
O=\frac{\sum_{p_{i} \in P}\sum_{p_{j} \in P}  \| p_{i}-p_{j}\|^{2})}{2 \cdot|P|}
\end{equation}
where $P$ represents the set of current selection of placement points and existing vehicles. $O$ is used to measure the average distances between all points. To achieve multi-view visibility with as uniform distribution as possible, we define the ultimate optimization goal as :
\begin{equation*}
\begin{split}
&\max_{P} \left[\sum_{p_{i }\in P} V_{i}+\frac{\sum_{p_{i} \in P}\sum_{p_{j} \in P}  \| p_{i}-p_{j}\|^{2})}{2 \cdot|P|}\right] 
\end{split}
\end{equation*}

\textbf{Rule-based checker.} After determining the sampling points, we need to check whether these points are placeable under the distribution conditions. The placement of new objects must satisfy two main conditions: 1) In 3D space, there should be no overlapping with existing simulated objects or real objects; 2) The object should be visible from at least one camera perspective when projected across multiple camera views. The first condition ensures that the simulated scene complies with physical laws, such as preventing objects from collisions. The second condition ensures that the inserted object is visible in the rendered image. Each insertion requires checking these two conditions to determine the final feasible placement location, enabling adaptive object placement.

\begin{figure*}[!htp]
\centering

\captionsetup[subfloat]{labelsep=none,format=plain,labelformat=empty}

        \rotatebox{90}{\scriptsize{~~~~~~~~~~~}}
	\subfloat[BEV-origin]{\includegraphics[width = 0.24\textwidth]{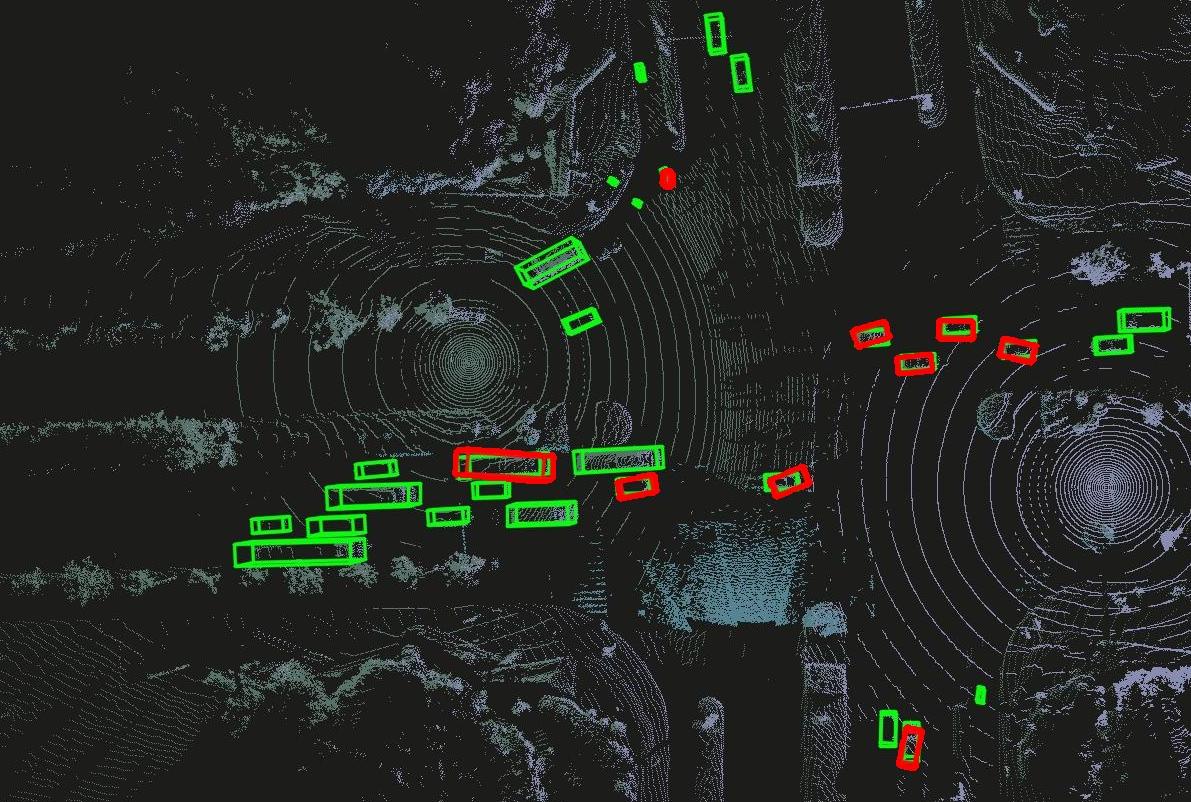}}
	\hfill
	\subfloat[BEV-sim]{\includegraphics[width = 0.24\textwidth]{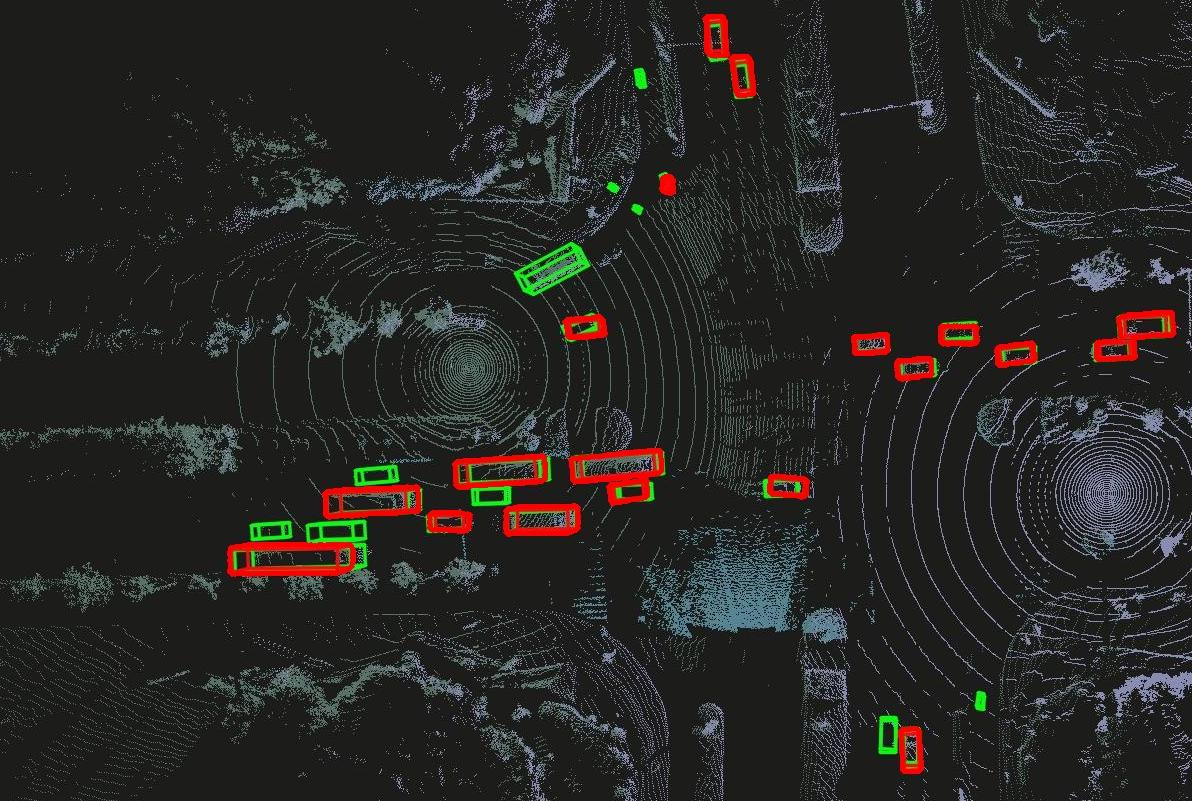}}
	\hfill
	\subfloat[3D-origin]{\includegraphics[width = 0.24\textwidth]{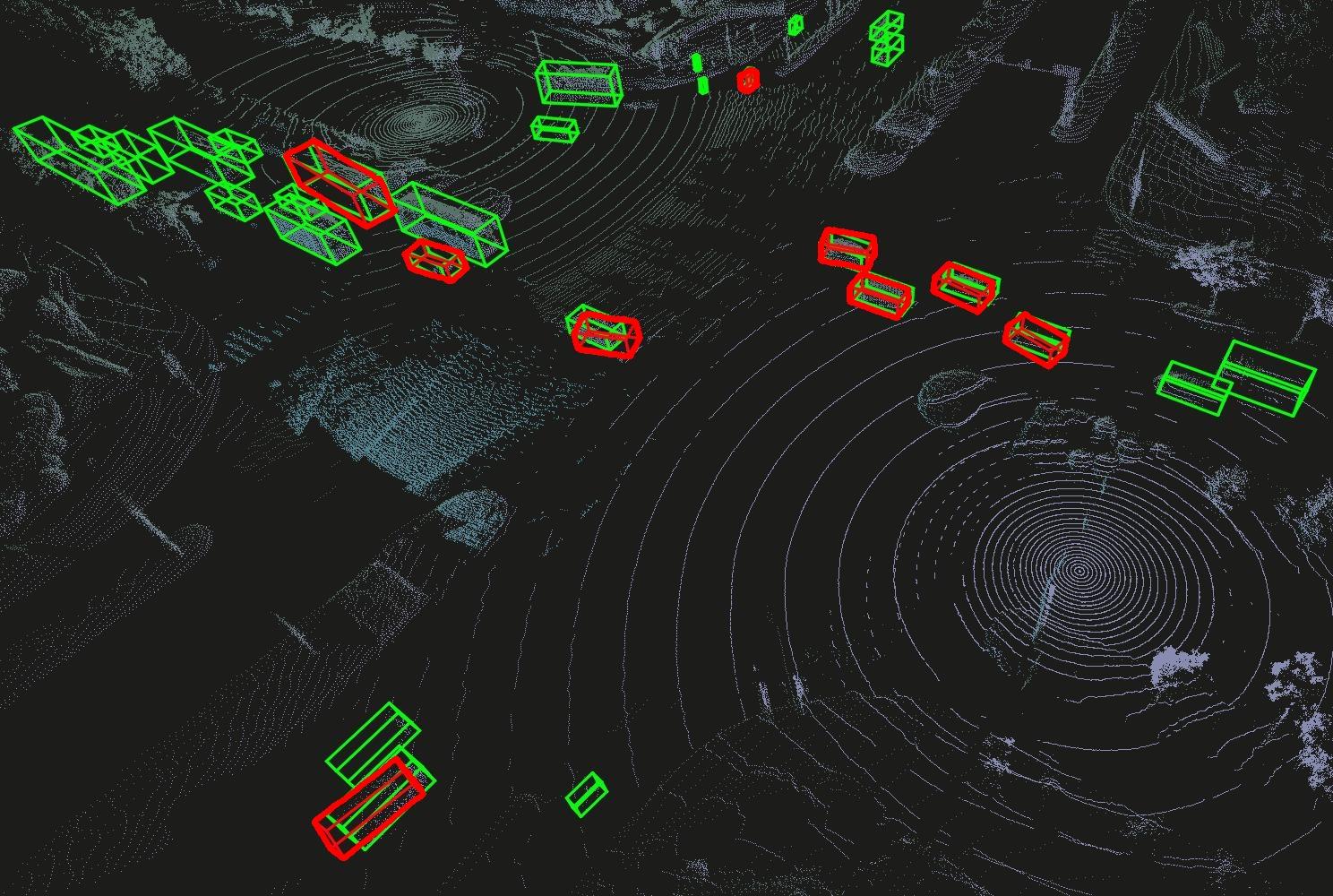}} 
    \hfill
	\subfloat[3D-sim]{\includegraphics[width = 0.24\textwidth]{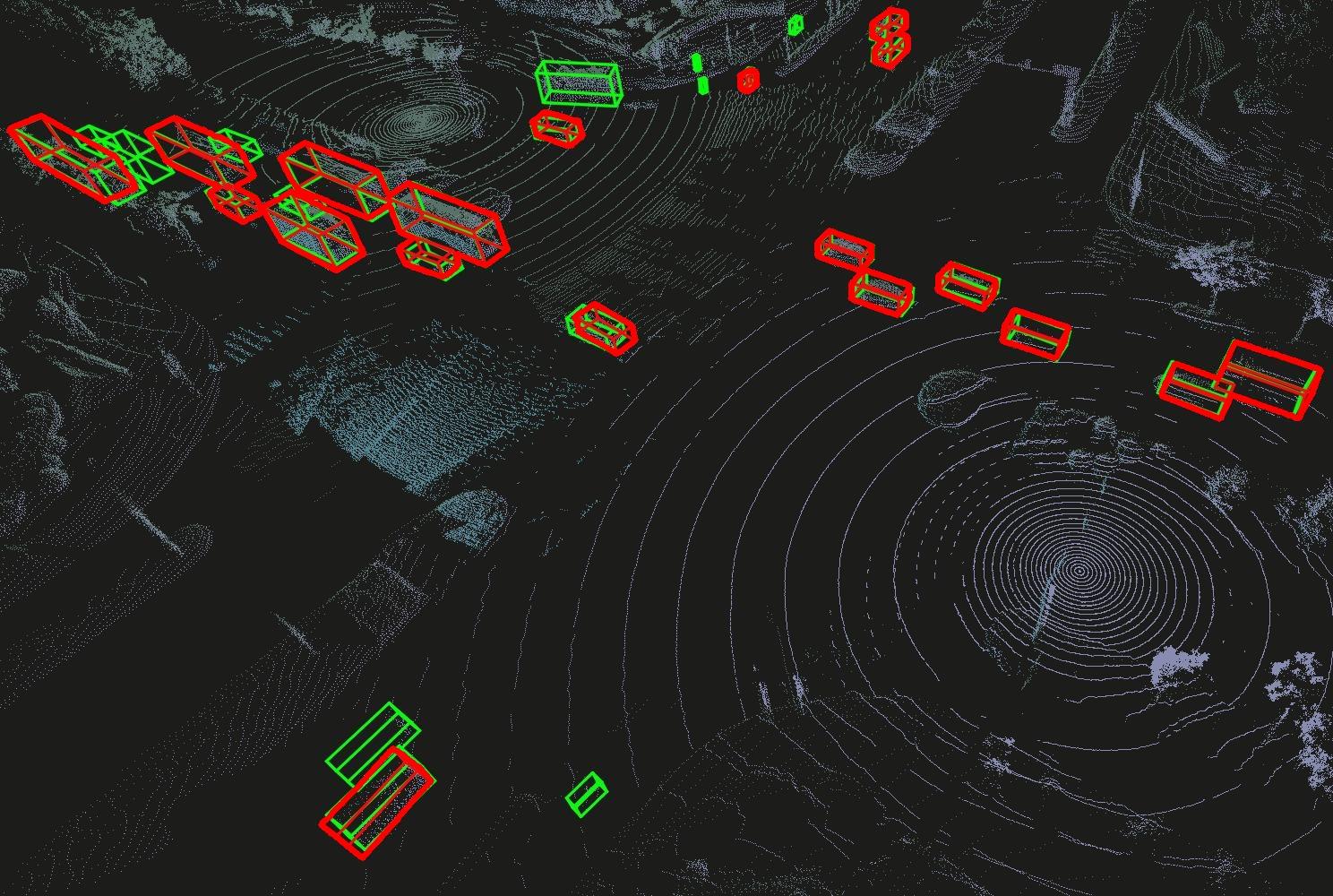}}  \\
\caption{The model trained with simulated data boosts vehicle detection. \textcolor{green}{Green} and \textcolor{red}{red} boxes denote GT and detection respectively. }
\label{fig:detection capabilities.}
\vspace{-0.6cm}
\end{figure*}

\subsection{DepthSAM}
\label{sec:depthSAM}
\hspace{1em} After modeling the camera and 3D foreground objects, we render them onto a 2D image while maintaining their 3D spatial relationships. To achieve this, we require a highly generalizable depth estimation method. However, existing approaches often struggle with ground and sky regions and lack absolute depth information. We address this by segmenting foreground objects, extracting their depth, and calibrating it with point cloud data to obtain accurate foreground depth for rendering while preserving multi-view consistency.

Given an image $I\in\mathbb{R}^{H \times W \times 3}$,  we can perform segmentation~\cite{mobilesam} to obtain the segmentation set $S(I)=\left\{S_{0}, S_{1}, \ldots S_{m}\right\}$, $m$ represents the number of instances segmented in the image. Using the annotation information, we can determine the center points of each foreground object $C_{\text {bbox }}=\left\{c_{0}, c_{1}, \ldots c_{n}\right\}$, thus obtaining the foreground segmentation as follows:
\vspace{-0.2cm}
 \begin{align}
S_{\text{foreground}} &= \left\{S_{i} \mid S_{i} \subseteq  S(I), \ N\left(S_{i}\right) < \frac{H \cdot W}{4} \right. \nonumber \\
&\quad \left. \text{and} \ \exists  \ c_{j} \in C_{\text{bbox}}, \ \text{s.t. } c_{j} \in S_{i} \right\}
\end{align}

    \hspace{1em}The number of points in \( S_{i} \), denoted as \( N(S_{i}) \), should satisfy the condition $N(s_{i}) < \frac{H \cdot W}{4}$
to filter the ground and sky. According to our statistics, the size of foreground objects typically does not exceed \( \frac{H \cdot W}{4} \). Then we can get mask $M\in\mathbb{R}^{H \times W}$ as follows:
\begin{equation}
M(u,v) =
\begin{cases}
1, & \text{if } (u,v) \in S_{\text{foreground}} \\
0, & \text{otherwise}
\end{cases}
\end{equation}

We denotes $D_{\text{rel}} \in\mathbb{R}^{H \times W}$ represents the relative depth predicted by DepthAnything~\cite{depthanything} and $Z \in\mathbb{R}^{H \times W}$ represents the projected depth map. The projected depth is obtained from a 3D point cloud $P_{3d} \in \mathbb{R}^{N \times 3}$ , where N is the number of points. The point cloud is projected into the camera coordinate system using $P_{\text{cam}} = \mathcal{M}_{\text{l2c}} \cdot P_{3d}$. The depth value $Z(u,v)$ satisfying $Z(u,v) \cdot (u,v,1)^T = \mathcal{K} \cdot P_{cam}$ for pixel $(u,v)$. We solve for the calibration parameters a and b using the following optimization problem:

\begin{equation}
\min_{a, b} \sum_{\substack{(u, v) \in P_{2d}}} \left\| a \cdot D_{rel}(u,v) + b - Z(u, v) \right\|^2
\end{equation}
where $P_{2d}$ is the set of 2D pixels in the image, $a$ and $b$ are the scaling factor and offset, respectively. They transform the relative depth $D_{rel}$ into absolute depth via $a \cdot D_{rel}(u,v) + b$. The calibrated foreground depth $D_{\text{final}}$ is defined as:
\begin{equation}
D_{\text{final}} =
\begin{cases}
a \cdot D_{\text{rel}}(u, v) + b, & M(u, v) = 1 \\
+\infty, & M(u, v) = 0
\end{cases}
\end{equation}
\hspace{1em} DepthSAM extracts pixel-level depth for foreground objects, ensuring correct occlusion and depth order during 3D-to-2D rendering, which guarantees multi-view consistency.

\subsection{Scalable Post-Processing Toolkit}
\label{sec: Scalable Post-Processing Toolkit}
\hspace{1em}To elevate the diversity of simulation scenarios and embrace a future-oriented approach, we design a scalable post-processing toolkit for RoCo-Sim. This toolkit provides a modular interface that seamlessly integrates a wide array of innovative image processing tools, including style transfer, scene transformation, and more. By doing so, it not only significantly enhances the diversity of the generated data but also paves the way for continuous expansion, allowing the incorporation of more sophisticated post-processing techniques to enrich the simulation environment.

\section{Experiment}

\begin{table*}[ht]
\centering
\vspace{-0.2cm}
\caption{RoCo-Sim greatly enhances detection performance, with the improvements brought by simulation even surpassing those achieved by the optimized model BEVSpread.}
\setlength{\tabcolsep}{1.2mm} 
\resizebox{0.8\textwidth}{!}{
\begin{tabular}{cccccccccccc}
   \toprule
   \multicolumn{1}{c}{\multirow{2}{*}{\textbf{Backbone}}}   &\multicolumn{1}{c}{\multirow{2}{*}{\textbf{Roco-Sim}}}  &\multicolumn{5}{c}{\textbf{Rcooper-117} } &\multicolumn{5}{c}{\textbf{Rcooper-136}}\\
   \cmidrule(r){3-7}                 \cmidrule(r){8-12} 
     \multicolumn{1}{c}{}& \multicolumn{1}{c}{} &origin/aug& bev@0.5 & 3D@0.5 & bev@0.7 & 3D@0.7 &origin/aug  & bev@0.5 & 3D@0.5 & bev@0.7 & 3D@0.7 \\
        \midrule 
     \multicolumn{1}{c}{\multirow{2}{*}{BEVHeight}}&$\times$ & 5120/0 & 65.3233 & 56.1948 & 46.9456 & 31.8694 & 1632/0 & 22.6618 & 14.8236 & 13.6882 &7.2549 \\
     \multicolumn{1}{c}{} & \checkmark & 5120/2400 & \textbf{77.6463} & \textbf{71.5983} & \underline{50.6319} & \textbf{42.0199} & 1632/4500  & \textbf{40.3816} & \underline{33.1069} & \textbf{29.9080} & \underline{16.6990} \\
     \cmidrule(r){1-2}  \cmidrule(r){3-7}                 \cmidrule(r){8-12} 
     \multicolumn{1}{c}{\multirow{2}{*}{BEVSpread}}&$\times$ &5120/0 &66.3986 & 57.4217 & 48.2224 & 34.5299 & 1632/0 &36.5752 &
     30.0382& 21.5373 & 13.2292\\


     
     \multicolumn{1}{c}{} & \checkmark & 5120/2400&\underline{75.5791} & \underline{69.3342}& \textbf{56.1572}&\underline{40.6480} &1632/4500 & \underline{40.1567}& \textbf{36.0443} & \underline{29.6339}& \textbf{19.0902}\\

        \bottomrule
        \toprule

   \multicolumn{1}{c}{\multirow{2}{*}{\textbf{Backbone}}}   &\multicolumn{1}{c}{\multirow{2}{*}{\textbf{Roco-Sim}}}  &\multicolumn{5}{c}{\textbf{Rcooper-117+Rcooper-136}}  &\multicolumn{5}{c}{\textbf{TUM}}\\
   \cmidrule(r){3-7}                 \cmidrule(r){8-12} 
     \multicolumn{1}{c}{}& \multicolumn{1}{c}{} &origin/aug& bev@0.5 & 3D@0.5 & bev@0.7 & 3D@0.7 &origin/aug  & bev@0.5 & 3D@0.5 & bev@0.7 & 3D@0.7 \\
     \midrule
          \multicolumn{1}{c}{\multirow{2}{*}{BEVHeight}}&$\times$ & 6752/0 & 36.1190 & 21.5442 & 19.9593 &10.8835& 2400/0 & 54.8044 &45.4383 & 28.3148 &13.1911 \\
     \multicolumn{1}{c}{} & \checkmark & 6752/9000 & \underline{54.6708} & 37.4621 & \underline{36.6740} & 15.0132 & 2400/2400 & \underline{57.0037} & 43.8220 &\underline{35.4583} &\textbf{24.1557}\\
    \cmidrule(r){1-2}  \cmidrule(r){3-7}                 \cmidrule(r){8-12} 
     \multicolumn{1}{c}{\multirow{2}{*}{BEVSpread}}&$\times$ &6752/0 &53.5341 & \underline{43.8613} & 34.7345 & \underline{20.1353}& 2400/0 &52.8383 &
     43.3036& 32.2100 & 19.8507\\


     
     \multicolumn{1}{c}{} & \checkmark & 6752/9000&\textbf{54.9626} & \textbf{48.9011}& \textbf{38.8669}&\textbf{26.7668} &2400/2400 & \textbf{59.3968}& \textbf{48.5001} & \textbf{37.0695}& \underline{23.0567}\\

     \bottomrule
\end{tabular}
}

\label{tab:RoCo-Sim}

\vspace{-0.4cm}
    
    
\end{table*}

\hspace{1em}In this section, we introduce our experimental setting and evaluate the effectiveness of RoCo-Sim on two roadside camera-only detection models. All experiments are conducted using the optimized extrinsic parameters obtained from Eq. \ref{equation3} by default. Results show RoCo-Sim boosts performance in both cases, which proves RoCo-Sim's value. We report the average precision at 40 recall points (\( AP_{3D|R40} \))~\cite{kitti} for 3D bounding boxes and assess the improvements in network capability driven by simulated data using the KITTI evaluation metrics.

\subsection{Datasets}
\hspace{1em}We use two of the latest  road-side collaborative perception datasets: Rcooper-Intersection~\cite{rcooper} and TUMTraf-V2X~\cite{tum}. We conduct simulations across three traffic scenarios in both datasets to evaluate the impact of simulation on improving the performance of 3D object detection tasks.\par
\textbf{Rcooper-Intersection.} We focus specifically on intersection scenarios due to their higher visual information overlap, which facilitates collaborative perception. The Rcooper-Intersection dataset comprises two traffic scenarios: 117-118-120-119 and 136-137-138-139 (hereafter referred to as 117 and 136, respectively). \par
\textbf{TUMTraf-V2X.} To assess the generalization improvements from simulation, we avoid the official data splits, where target information in training and validation sets is nearly identical. Instead, we apply temporal partitioning, sorting data chronologically into sequences of 40 images and using the last two sequences (13th and 14th) for validation. This ensures a temporal separation between training and validation sets, preventing misleading performance gains from overfitting.
\subsection{Implementation Details}
\hspace{1em}
\textbf{Perception model.} We conduct experiments using BEVHeight~\cite{yang2023bevheight} and BEVSpread~\cite{wang2024bevspread}, setting the perception range to 0–150 meters with ResNet-101 as the image backbone. The grid size is fixed at 0.4 meters, and the initial learning rate is set to 2e-4. All experiments are conducted on two RTX-4090 GPUs, with models trained for 40 epochs, ensuring proper convergence.\par
\textbf{Simulation baseline.} Currently, no simulation methods are designed for roadside scenarios. Existing vehicle-side approaches rely on multi-view cameras or 3D reconstruction techniques like NeRF and 3DGS, which are unsuitable for fixed roadside cameras. Additionally, many simulation methods that generate 3D bounding boxes for downstream tasks require 3D layout inputs, which roadside datasets lack, limiting their applicability.

\begin{figure}
    \centering
    \captionsetup[subfloat]{labelsep=none,format=plain,labelformat=empty}
            \subfloat[(a)]{\includegraphics[width=0.235\textwidth]{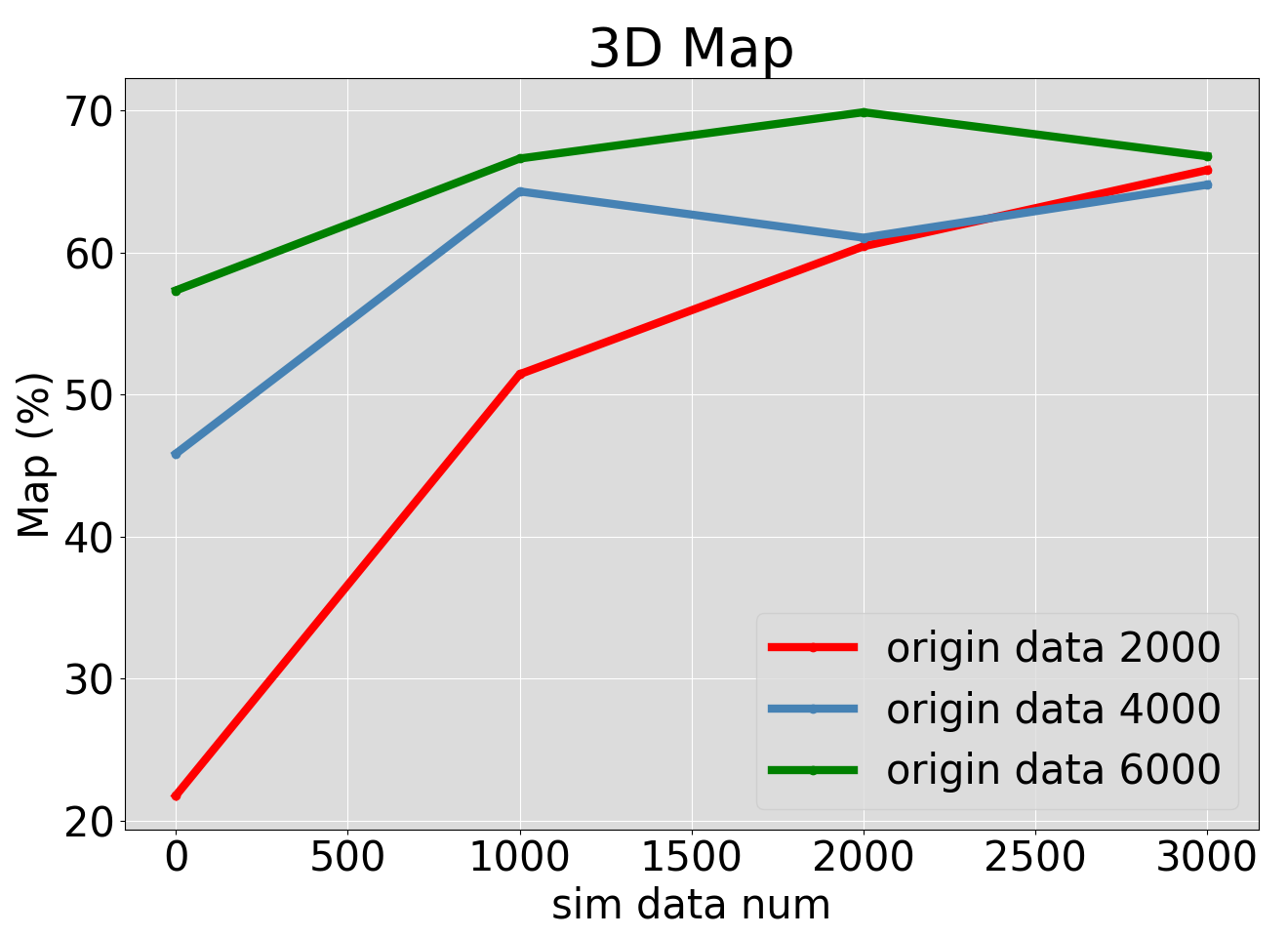}}
    \hfill
        \subfloat[(b)]{\includegraphics[width=0.235\textwidth]{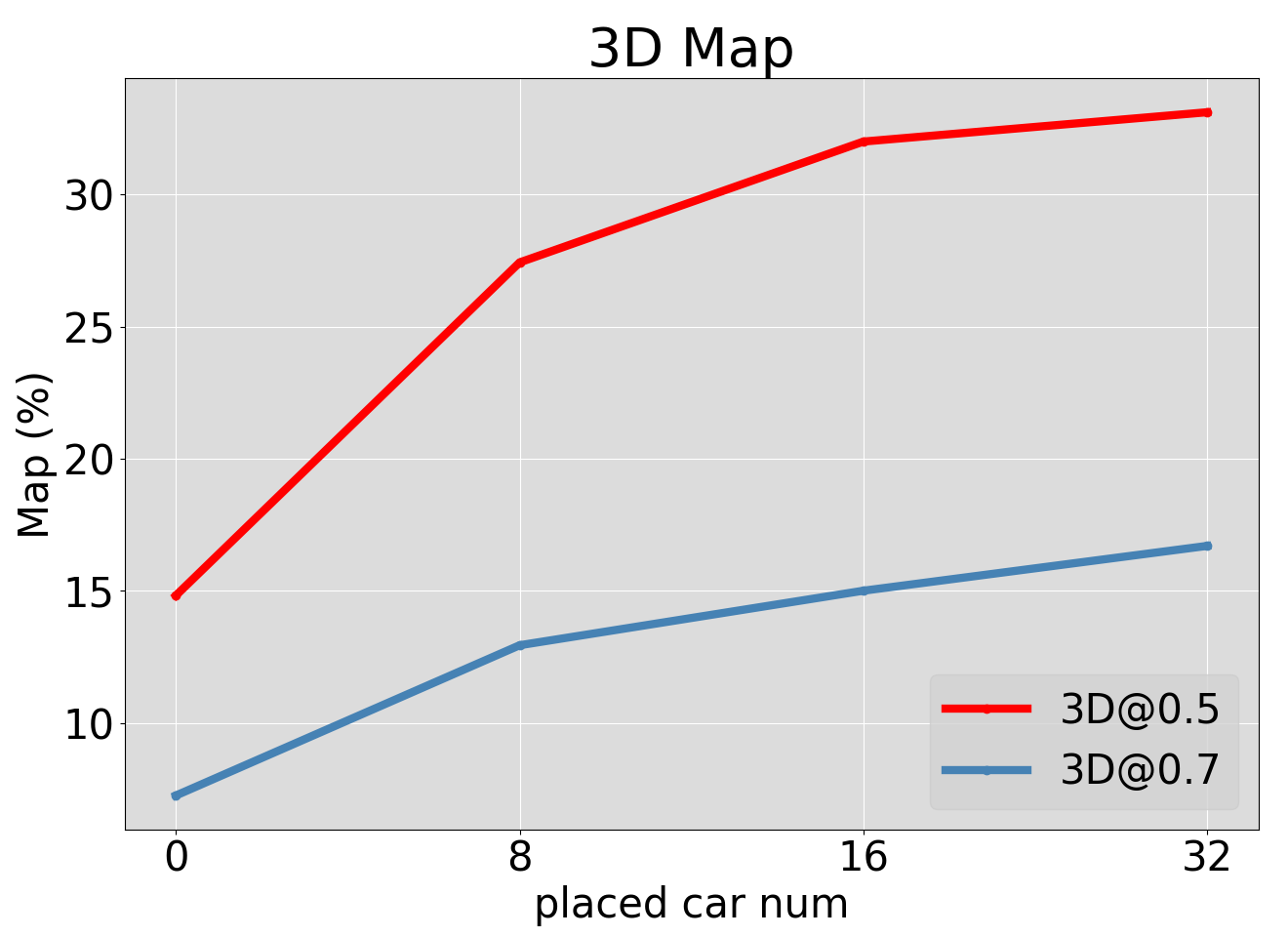}}
        \vspace{-0.3cm}
    \caption{(a) More simulations improve performance, allowing small-scale original data to achieve results comparable to larger datasets. (b) As the number of simulated vehicles increases, model performance improves.}
    \label{fig:more simulations}
    \vspace{-0.5cm}
\end{figure}

\begin{figure}
    \centering
    
    \includegraphics[width=1\linewidth]{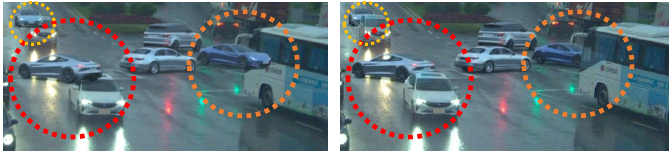}
    \vspace{-0.5cm}
    \caption*{w/o DepthSAM~~~~~~~~~~~~~~~~~~~~~ DepthSAM}
    \vspace{-0.3cm}
    \caption{Without DepthSAM, simulated vehicles may overlap with existing ones, disrupting spatial relationships.}
    \label{fig:depthsam}
    \vspace{-0.7cm}
\end{figure}



\subsection{Simulation Results}
\hspace{1em}
From Fig.\ref{fig:Visual Results of the Simulation}, it can be seen that RoCo-Sim demonstrates outstanding simulation performance, addressing various data issues and overcoming the challenges of roadside simulation. Fig.\ref{fig:Visual Results of the Simulation} (a) highlights the critical impact of extrinsic parameter optimization on data annotation, effectively correcting a significant number of projection errors in the annotations. Fig.\ref{fig:Visual Results of the Simulation} (b) illustrates that simulated vehicles seem to float before optimization due to inaccurate extrinsic parameters affecting camera modeling. After optimization, vehicles are more realistically aligned with the ground, as correct extrinsic parameters ensure precise rendering of simulated objects. 
Fig.\ref{fig:Visual Results of the Simulation} (c) shows that RoCo-Sim correctly determines the relative depth order between simulated and real vehicles, further validating the effectiveness of DepthSAM. Fig. \ref{fig:Visual Results of the Simulation} (d) illustrates RoCo-Sim's multi-view consistency for simulated vehicles. With MOAS and DepthSAM, vehicle placement is sensible, projections are accurate, and there are correct occlusion relationships between objects across views. Fig.\ref{fig:Visual Results of the Simulation} (e) shows very realistic simulations with four vehicles added to a real scene, making the synthetic parts hard to spot. It also showcases  RoCo-Sim's potential to generate diverse data, such as in rainy and night conditions, using the Scalable Post-Processing Toolkit. Fig. \ref{fig:Visual Results of the Simulation} (f) illustrates the dynamic foreground editing.

\begin{table*}[ht]
\centering
\caption{Incorporating simulated data generated by RoCo-sim improves late fusion performance in both models.
}
\setlength{\tabcolsep}{1.2mm} 
\resizebox{0.8\textwidth}{!}{
\begin{tabular}{cccccccccccc}
   \toprule
   \multicolumn{1}{c}{\multirow{2}{*}{\textbf{Backbone}}}   &\multicolumn{1}{c}{\multirow{2}{*}{\textbf{Roco-Sim}}}  &\multicolumn{5}{c}{\textbf{Rcooper-117} } &\multicolumn{5}{c}{\textbf{Rcooper-136}}\\
   \cmidrule(r){3-7}                 \cmidrule(r){8-12} 
     \multicolumn{1}{c}{}& \multicolumn{1}{c}{} &origin/aug& bev@0.5 & 3D@0.5 & bev@0.7 & 3D@0.7 &origin/aug  & bev@0.5 & 3D@0.5 & bev@0.7 & 3D@0.7 \\
        \midrule 
     \multicolumn{1}{c}{\multirow{2}{*}{BEVHeight}}
     &$\times$ & 5120/0 & 12.8650 & 7.3672 & 7.1578 & 3.7598 & 1632/0 & 14.3390 & 9.0959 & 8.1261 &4.0682 \\
     \multicolumn{1}{c}{} & \checkmark & 5120/2400 & \underline{16.6556} & \underline{16.2424} & \underline{13.0317} & \underline{9.3716} & 1632/4500  & 15.2574 & 9.0587 & \textbf{21.1498} & \textbf{17.6615} \\
     \cmidrule(r){1-2}  \cmidrule(r){3-7}                 \cmidrule(r){8-12} 
     \multicolumn{1}{c}{\multirow{2}{*}{BEVSpread}}&$\times$ &5120/0 &16.5150 & 13.6578& 10.7850 & 7.7328 & 1632/0 &\underline{20.5816} &
     \underline{17.0001}& 13.2643 & 7.7018\\
     \multicolumn{1}{c}{} & \checkmark & 5120/2400&\textbf{16.9561} & \textbf{16.3672}& \textbf{13.1573}&\textbf{10.0375} &1632/2400 & \textbf{21.3933}& \textbf{18.3458} & \underline{15.0851}& \underline{11.0922}\\
        \bottomrule
        \toprule
    \multicolumn{1}{c}{\multirow{2}{*}{\textbf{Backbone}}}   &\multicolumn{1}{c}{\multirow{2}{*}{\textbf{Roco-Sim}}}  &\multicolumn{5}{c}{\textbf{Rcooper-117+Rcooper-136}}  &\multicolumn{5}{c}{\textbf{TUM}}\\
   \cmidrule(r){3-7}                 \cmidrule(r){8-12} 
     \multicolumn{1}{c}{}& \multicolumn{1}{c}{} &origin/aug& bev@0.5 & 3D@0.5 & bev@0.7 & 3D@0.7 &origin/aug  & bev@0.5 & 3D@0.5 & bev@0.7 & 3D@0.7 \\
     \midrule
          \multicolumn{1}{c}{\multirow{2}{*}{BEVHeight}}
          &$\times$ & 6752/0 & 14.7159 & 8.7044 & 8.3179 &4.0130& 2400/0 & \underline{42.7962} &\textbf{33.1987} & 18.6450 &8.9220 \\
     \multicolumn{1}{c}{} & \checkmark & 6752/9000 & \textbf{21.4400} & \textbf{17.8329} & \underline{14.7487} & \underline{7.6758} & 2400/2400 & 42.4077 & \underline{33.1671} &\textbf{25.1554} &\textbf{16.0500}\\
     \cmidrule(r){1-2}  \cmidrule(r){3-7}                 \cmidrule(r){8-12} 
     \multicolumn{1}{c}{\multirow{2}{*}{BEVSpread}}&$\times$ &6752/0 &13.5685 & 12.3687 & 9.7065 & 6.8483 & 2400/0 &40.7025 &
     31.6073& 21.4561 &\underline{14.0298}\\
     \multicolumn{1}{c}{} & \checkmark & 6752/9000&\underline{19.0796} & \underline{16.3904}& \textbf{14.8512}&\textbf{9.3302} &2400/2400 & \textbf{44.9000}& 32.1661 & \underline{22.8040}& 13.2693\\

     \bottomrule

\end{tabular}
}

\label{tab:late fusion}

    \vspace{-0.4cm}
\end{table*}

\begin{table}[ht]
    \vspace{-0.2cm}
    \caption{Camera Extrinsic Optimization improves detection performance}
    \centering
    \vspace{-0.2cm}
    \resizebox{\columnwidth}{!}{
        \begin{tabular}{cccccc}
            \toprule
            \textbf{Dataset} &\textbf{Calib} & bev@0.5 & 3D@0.5 & bev@0.7 & 3D@0.7\\
            \midrule
           \multicolumn{1}{c}{\multirow{2}{*}{117}}  &$\times$ &39.6378& 33.0679 & 25.3091 & 19.6054 \\
            & \checkmark &65.3233 & 56.1948 & 46.9456 & 31.8694 \\
           \cmidrule(r){1-2}  \cmidrule(r){3-6}   
           \multicolumn{1}{c}{\multirow{2}{*}{136}}  &$\times$& 18.2512& 12.8897 & 10.8723 & 7.9102 \\
            &\checkmark&
           22.6618 & 14.8236 & 13.6882 & 7.2549\\
           \cmidrule(r){1-2}  \cmidrule(r){3-6}
           \multicolumn{1}{c}{\multirow{2}{*}{117-136}}  &$\times$& 32.2033& 19.6822 & 17.9436 & 6.2302 \\
            &\checkmark&
           36.1190 & 21.5442 & 19.9593 & 10.8835 \\
            \bottomrule
        \end{tabular}
    }
    \label{tab:calib optimization}
\vspace{-0.2cm}
\end{table}

\begin{table}[ht]
    \caption{Scorer increases the number and diversity of vehicles rendered onto images, Checker prevents object overlap, and DepthSAM ensures correct depth relationships during rendering. All contribute to better simulation data generation and improved model detection performance.}
    \centering
    \vspace{-0.2cm}
    \resizebox{0.7\textwidth}{!}{
    \begin{minipage}{\textwidth}
        \begin{tabular}{ccccccc}
        \toprule
          \makebox[0.1\textwidth][c]{DepthSAM}\quad &\makebox[0.05\textwidth][c]{Scorer} &   \makebox[0.06\textwidth][c]{Checker}  &  bev@0.5 & 3D@0.5 & bev@0.7 & 3D@0.7 \\
            \midrule
             \checkmark & &   & 31.2110 & 19.6172 &18.7138 & 9.6544 \\
             \checkmark&  & \checkmark & 38.2746 & 23.3362 & 23.0323 & 9.0517 \\
            \checkmark& \checkmark &  &33.1691 & 21.4200 & 22.9378 & 9.4178 \\
            & \checkmark &\checkmark  & 37.2985 & 16.9951 &21.5125 & 5.5527\\
            \checkmark & \checkmark &\checkmark & \textbf{39.5758} & \textbf{31.9978} & \textbf{25.7230} & \textbf{15.0032}\\
        \bottomrule
        \end{tabular}
        
    \end{minipage}
    }
    \label{tab:scorer pp-checker}
    \vspace{-0.5cm}
\end{table}

\subsection{Perception Evaluation}
\vspace{-0.1cm}
\hspace{1em}\textbf{RoCo-Sim significantly enhances both roadside perception and late fusion in roadside cooperative perception.} 
Performance results are shown in Fig.\ref{fig:bar_plot}, for BEVHeight, we see major improvements: AP3D 50 rises by 27.4\%, 143.2\%, 127.0\% for Rcooper in scenarios 117, 136, and 117-136. More numerical results are shown in Tab. \ref{tab:RoCo-Sim}, demonstrating simulation data's strong impact. Tab. \ref{tab:late fusion} underscores late fusion's role in enhancing perception by combining multi-view data beyond single-camera capabilities.
In Tab. \ref{tab:RoCo-Sim}, labels are limited to 3D annotations visible in the image, while Tab. \ref{tab:late fusion} includes all 3D scene objects, even those outside the camera view.
Fig. \ref{fig:detection capabilities.} shows that models trained with additional simulation data can detect more objects in the late fusion stage. Moreover, the positions and orientations of the bounding boxes are more accurate.

\textbf{The performance improvement through RoCo-Sim far exceeds that of algorithm enhancements.} Comparing with the newer BEVSpread~\cite{wang2024bevspread}, we find that BEVHeight~\cite{yang2023bevheight} trained on simulation data outperforms BEVSpread trained on original data. This finding highlights that simulated data not only enhances performance but can surpass gains from algorithmic improvements.

\textbf{Impact of simulation data scale and vehicles.}  We study how adding simulated data to different initial data sizes on the 117 dataset affects results, as shown in Fig. \ref{fig:more simulations} (a). Mixing just 2000 original images with lots of simulated data greatly improves network performance, matching that of networks trained on large real datasets. This shows simulated data can effectively supplement real data. We also look at detection performance changes on the 136 dataset as the number of simulated vehicles increased, as in Fig. \ref{fig:more simulations} (b). More vehicles mean better detection, suggesting we can use fewer simulations by adding more vehicles to get similar results, cutting training costs.


       

\vspace{0.2cm}
\subsection{Ablation studies}
\hspace{1em}We validate key components of RoCo-Sim. The scalable post-processing toolkit enhances style diversity for better generalization and real-world deployment, but in the ablation study, it is only used to maintain foreground-background style consistency.

\textbf{Camera Extrinsic Optimizer.} From Tab. \ref{tab:calib optimization}, we find that calibrating the camera extrinsics can significantly enhance model performance. For instance, in the case of 117, as shown in Fig. \ref{fig:Visual Results of the Simulation}, the extrinsics are quite inaccurate, resulting in a 62.55\% improvement in performance for AP70 after calibration. This is because camera extrinsics calibration aligns the images and labels, making it easier for the model to learn useful information from them.\par
\textbf{DepthSAM.} From Tab. \ref{tab:scorer pp-checker}, We find that without DepthSAM, performance is poor and may even decline in AP70. This is because, without DepthSAM, vehicles behind in the original image can obscure those in front during rendering, as show in Fig. \ref{fig:depthsam},  leading to a violation of the spatial relationships and resulting in decreased performance. \par
\textbf{MOAS.} From Tab. \ref{tab:scorer pp-checker}, we find that both the Scorer and Checker contribute to enhancing the quality of generated simulation data. The Scorer macroscopically regulates vehicle placement to allow for more visibility with the same number of vehicles, thus increasing the data scale. Meanwhile, the Checker ensures that the inserted vehicles do not conflict in 3D space and are visible in the projected image. We can also see that the impact of checker is greater than that of scorer. This is because without checker, the placed simulated vehicles would overlap, which is not possible in the real world, making it difficult for the model to learn.

\vspace{-0.8cm}

\section{Conclusion}
\vspace{0cm}
\hspace{1em}This work unleashes the power of simulation data to enhance roadside collaborative perception, demonstrating that simulation data, rather than model architecture, is the \textbf{true winner} in performance competition. The proposed RoCo-Sim is the first simulation framework specifically designed for roadside collaborative perception. RoCo-Sim generates a large scale of multi-view consistent simulation data from sparse fixed viewpoints, supports both foreground editing and full-scene style transformation, and can be rapidly deployed to any new road environment. We hope this work accelerates practical roadside collaborative perception development, ultimately enhancing driving safety.\par
\textbf{Limitation and future work.} 
Current 3D assets are limited in variety, and manually constructing them is costly. Therefore, it is important to explore new methods for generating diverse 3D assets. Besides, a trajectory generator that considers real-world traffic flow and regulations could further enhance the realism of virtual vehicle distributions. Additionally, we find that existing models experience a significant performance drop when applied to a new intersection. In the future, we hope that our simulation framework can help train detection models with stronger generalization capabilities.\par
{
    \small
    \bibliographystyle{ieeenat_fullname}
    \bibliography{main}
}

\end{document}


\subsection{DepthSAM vs. Existing Methods}
\quad Existing depth estimation methods can be broadly categorized into two types. The first is monocular depth estimation, which directly predicts pixel-wise depth from a single RGB image (e.g. Depth-Anything\cite{depthanything}). However, such methods can only estimate relative depth rather than true depth, making them unsuitable for direct deployment in downstream tasks. The second is depth completion (e.g. OMNI-DC\cite{omnidc}), which can obtain pixel-wise absolute depth by fusing sparse LiDAR point clouds with images. However, its performance heavily depends on the density and coverage of the point cloud.

Our proposed DepthSAM addresses these limitations by leveraging foreground object point clouds to calibrate the relative depth predicted by monocular depth estimators, thereby producing accurate pixel-level absolute foreground depth. This enables more precise foreground depth, which is crucial for maintaining depth consistency when rendering 3D objects into 2D images.

\begin{figure}[!h] 
\centering
\includegraphics[width=0.95\linewidth]{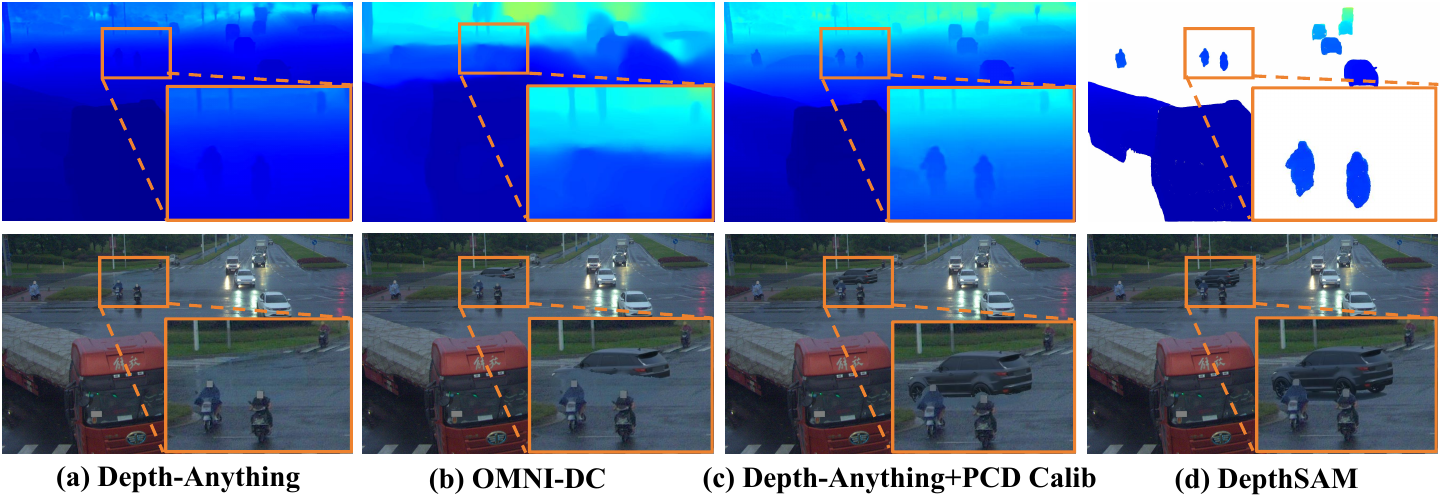}
\caption{Row1: The visual results of depth estimation using different methods; Row2: Comparison of rendering results using depth maps obtained from different depth estimation methods. DepthSAM provides the most accurate foreground depth, resulting in a more realistic relative positioning after rendering.}
\label{fig:depthsam comparison}
\end{figure}

As shown in Tab.\ref{tab:depthsam table}, DepthSAM achieves the lowest error on foreground objects depth estimation. Furthermore, Fig.\ref{fig:depthsam comparison} illustrates that accurate foreground pixel-level depth is critical for high-quality 3D-to-2D rendering. The depth maps generated by DepthSAM lead to the most visually consistent rendering results.

\setcounter{table}{0}
\begin{table}[!htbp]
    \centering
    \caption{Depth error using a random 30\% of foreground LiDAR points as ground truth; the rest used as input without training.}
    \label{tab:depthsam table}
    \vspace{-0.2cm}

    \resizebox{1\columnwidth}{!}{
        \begin{tabular}{cccccc}
            \toprule
            Methods & Depth-Anything & OMNI-DC & Depth-Anything+PCD Calib & DepthSAM \\
            \midrule
            MAE $\downarrow$ & 19.8900 & 8.7828 & 5.9703 & \textbf{1.0487} \\
            REL $\downarrow$ & 0.5044 & 0.3170 & 0.3806 & \textbf{0.0350} \\
            \bottomrule
        \end{tabular}
    }
\end{table}

\vspace{-0.8cm}
\vspace{-0.5cm}

\subsection{Multi-view Simulation Results of RoCo-Sim}
\begin{figure}[!htbp]
    \centering
    \includegraphics[width=0.95\linewidth]{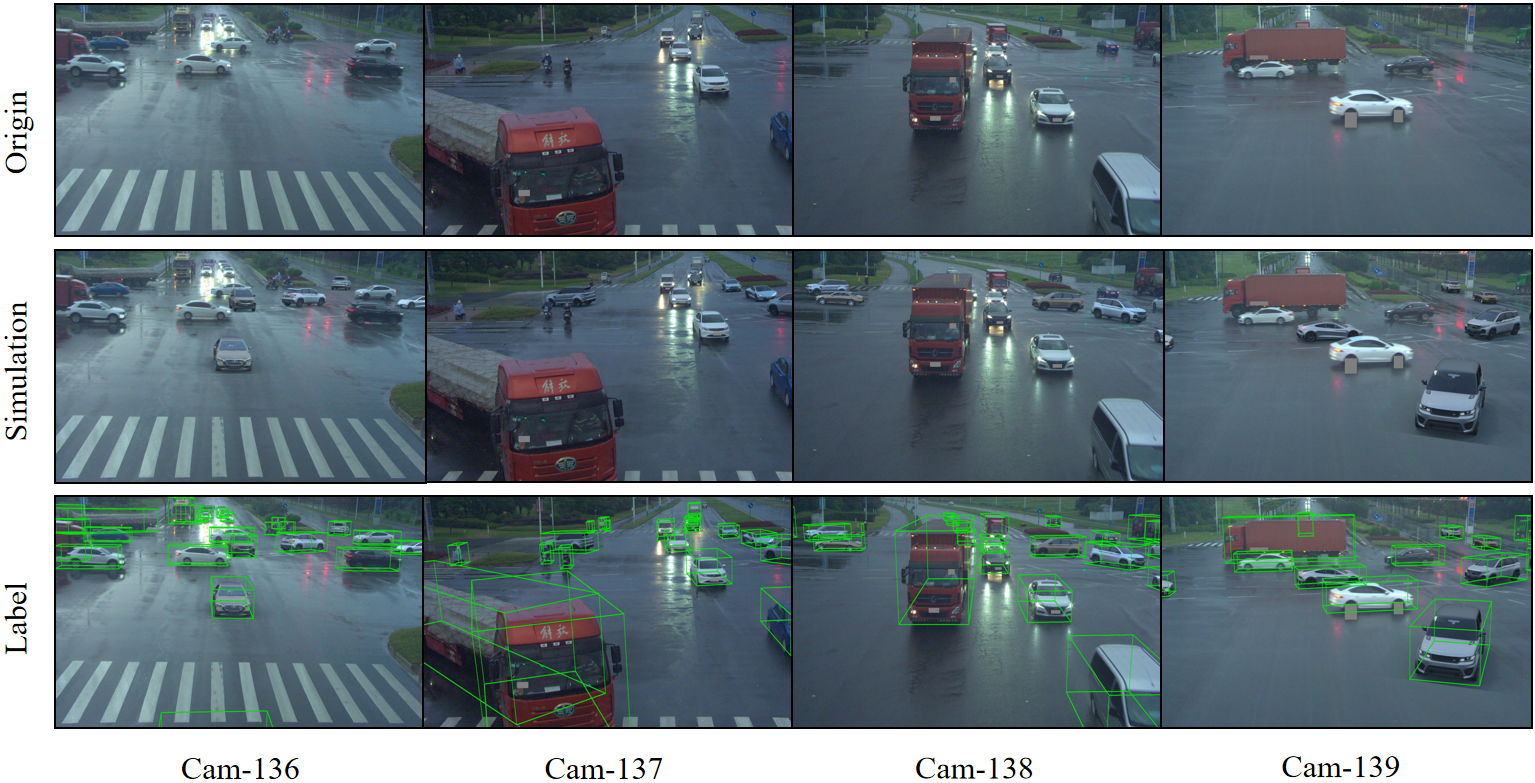}
    \caption{RoCo-Sim conducts multi-view simulation on Rcooper-136, and the simulation foreground has accurate 3D information.}
    \label{fig:136_sim}
\end{figure}

\quad Fig.\ref{fig:136_sim} shows the rendering results from multiple viewpoints at the 136 intersection. As illustrated, RoCo-Sim achieves realistic and consistent multi-view rendering, with accurate 3D information that can be leveraged for downstream tasks.

\subsection{Camera Extrinsic Optimization for Accurate 3D Alignment}
\quad The camera extrinsic optimization algorithm is evaluated on the 16-camera setup of RCooper. The results show that, after calibration, the projected 3D bounding boxes align more accurately with the vehicles, demonstrating the effectiveness and generalizability of the optimization method, as well as its applicability to roadside perception data.
\begin{figure*}[h]
    
    \centering
    \begin{subfigure}[b]{\textwidth}
        \includegraphics[width=0.95\textwidth]{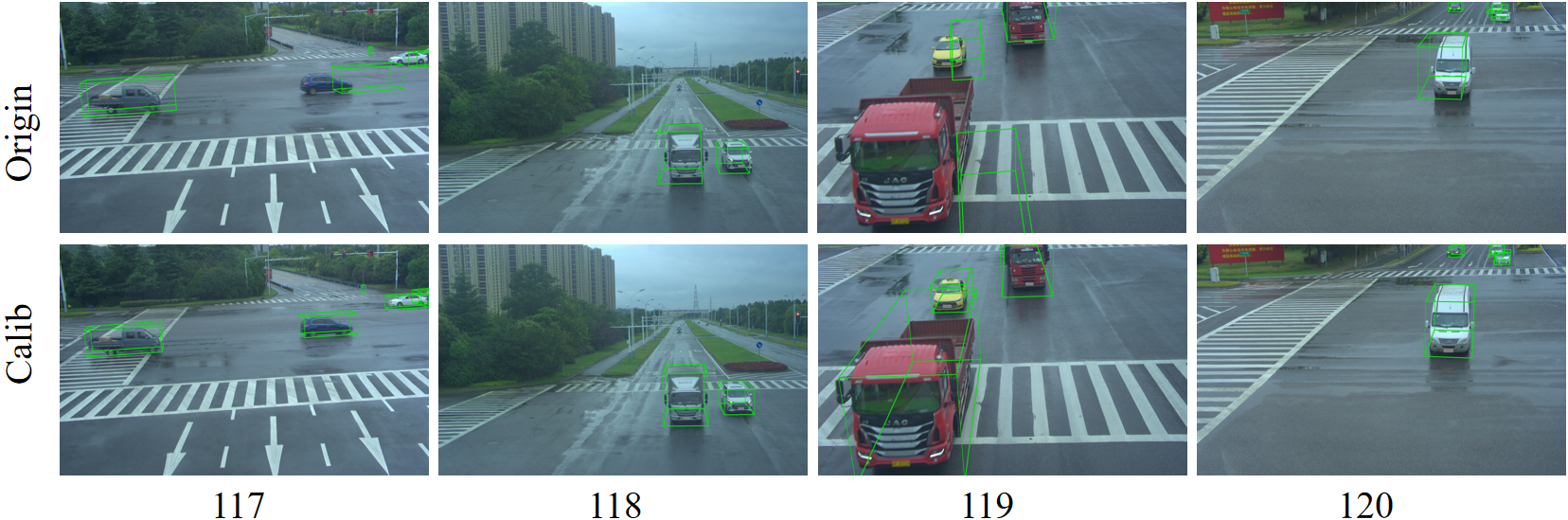}
        \label{fig:117}
    \end{subfigure}
    
    \begin{subfigure}[b]{\textwidth}
        \includegraphics[width=0.95\textwidth]{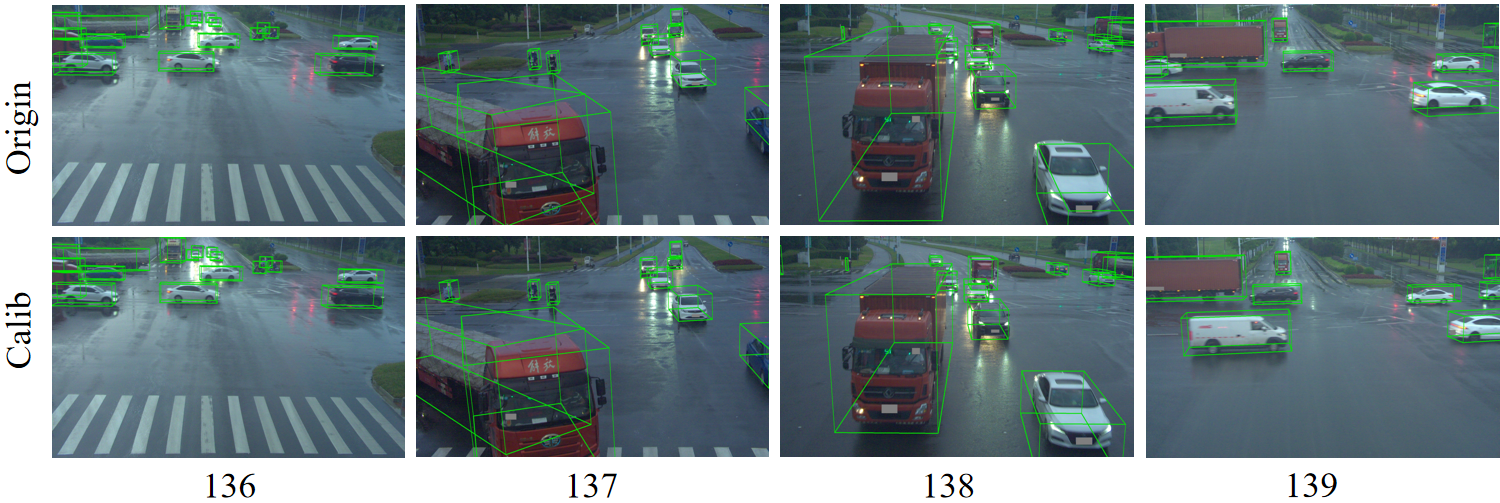}
        \label{fig:136}
    \end{subfigure}
    
    \begin{subfigure}[b]{\textwidth}
        \includegraphics[width=0.95\textwidth]{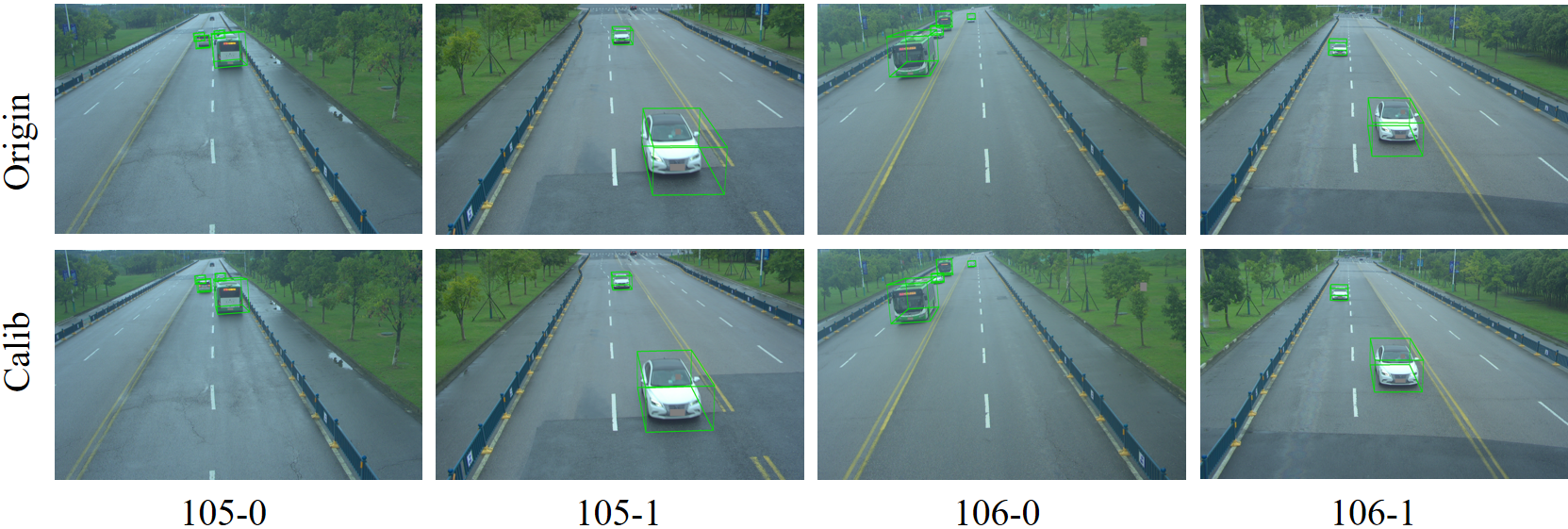}
        \label{fig:106}
    \end{subfigure}
    
    \begin{subfigure}[b]{\textwidth}
        \includegraphics[width=0.95\textwidth]{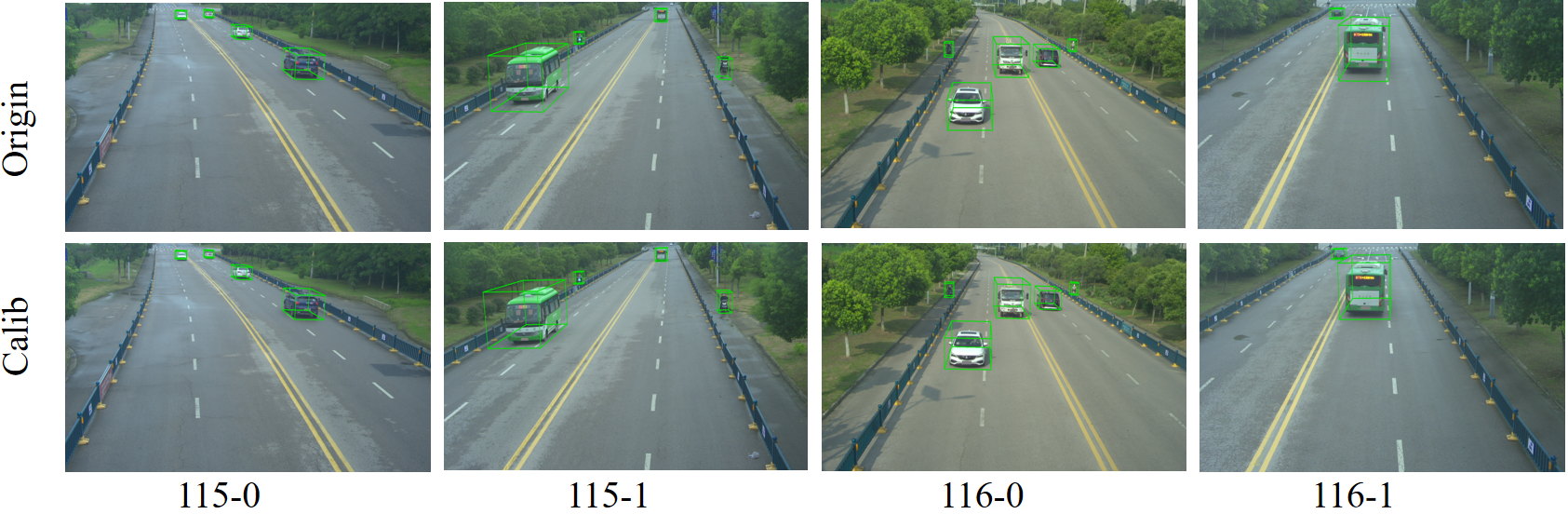}
        \label{fig:116}
    \end{subfigure}
    
    \caption{Camera Extrinsic Optimization results of 16 cameras of Rcooper.}
    \label{fig:camera_calibration}
\end{figure*}

\bibliographystyle{ieeenat_fullname}
\bibliography{supp}